\pdfoutput=1

\documentclass[11pt]{article}

\usepackage[final]{acl}

\usepackage{times}
\usepackage{latexsym}

\usepackage{booktabs}
\usepackage{tabularx}
\usepackage{xtab}
\usepackage{graphicx}
\usepackage{multirow}
\usepackage[T1]{fontenc}

\usepackage[utf8]{inputenc}

\usepackage{microtype}

\usepackage{subfig}

\usepackage{ulem}

\usepackage{inconsolata}

\usepackage{cleveref}

\usepackage[nolist]{acronym}

\usepackage{arydshln} %
\newcommand{\ie}{i.\,e., }
\newcommand{\eg}{e.\,g., }

\newcommand{\ds}{\mbox{\textsc{Deep Spectrum }}}

\newcommand{\deberta}{DeBERTaV3 }

\newcommand{\revision}[1] {{\color{black} #1}}

\title{Modeling Emotional Trajectories in Written Stories Utilizing Transformers and Weakly-Supervised Learning}

\author{Lukas Christ$^1$, Shahin Amiriparian$^2$, Manuel Milling$^2$, Ilhan Aslan$^3$, Bj\"orn W. Schuller$^{1,2,4}$ \\
         $^1$ EIHW, University of Augsburg, Germany $^2$CHI, TU Munich, Germany \\ $^3$ Device Software Lab, Huawei Technologies, Germany $^4$ GLAM, Imperial College London, UK\\
         \texttt{lukas1.christ@uni-a.de} \\}

\begin{document}
\maketitle
\begin{abstract}
Telling stories is an integral part of human communication which can evoke emotions and influence the affective states of the audience. Automatically modeling emotional trajectories in stories has thus attracted considerable scholarly interest. However, as most existing works have been limited to unsupervised dictionary-based approaches, there is no benchmark for this task. We address this gap by introducing continuous valence and arousal labels for an existing dataset of children's stories originally annotated with discrete emotion categories. We collect additional annotations for this data and map the categorical labels to the continuous valence and arousal space. For predicting the thus obtained emotionality signals, we fine-tune a DeBERTa model and improve upon this baseline via a weakly supervised learning approach. The best configuration achieves a Concordance Correlation Coefficient (CCC) of $.8221$ for valence and $.7125$ for arousal on the test set, demonstrating the efficacy of our proposed approach. A detailed analysis shows the extent to which the results vary depending on factors such as the author, the individual story, or the section within the story. In addition, we uncover the weaknesses of our approach by investigating examples that prove to be difficult to predict.
\end{abstract}

\section{Introduction}

Stories are central to literature, movies, and music, but also human dreams and memories~\cite{gottschall2012storytelling}. %
Storytelling has received widespread attention from various disciplines for many decades~\cite{polletta2011sociology},  %
\eg in the fields of psychology~\cite{sunderland2017using}, %
cognitive sciences~\cite{burke2015neuroaesthetics}, and history~\cite{palombini2017storytelling}.
A crucial aspect of stories is their emotionality, as stories typically evoke a range of different emotions in the listeners or readers, which also serves the purpose of keeping the audience interested~\cite{hogan2011affective}. 

Several efforts have been made to model emotionality in written stories computationally. However, these studies have often been constrained to dictionary-based methods~\cite{reagan2016emotional, somasundaran2020emotion}.
In addition, existing work often models emotions in stories on the sentence level only~\cite{agrawal2012unsupervised, batbaatar2019semantic} without taking into account surrounding sentences, missing out on important contextual information. In this study, we address the aforementioned issues by employing a pretrained \ac{LLM} to predict emotionality in stories automatically. 
In combination with an emotional \ac{TTS} system~\cite{triantafyllopoulos2022overview, amiriparian2023guest}, our system could serve naturalistic human-machine interaction, educational, and entertainment purposes~\cite{lugrin2010exploring}. For example, stories could be automatically read to children~\cite{eisenreich2014tale} by voice assistants.
Furthermore, the prediction of emotions in literary texts is of interest in the field of Digital Humanities~\cite{kim2018survey}, especially in Computational Narratology~\cite{mani2014computational, piper2021narrative}. 

We conduct our experiments on the children's story dataset created by~\citet{alm2008affect}. Specifically, our contributions are the following. First, we extend the annotations provided by~\citet{alm2008affect} and, subsequently, map the originally discrete emotion labels to the continuous valence and arousal~\cite{russell1980circumplex} space (cf.~\Cref{sec:data}). 
We then employ DeBERTaV3 in combination with a weakly-supervised learning step to predict valence and arousal in the stories provided in the dataset (cf.~\Cref{sec:exp-setup}). To the best of our knowledge, our work is the first to model emotional trajectories in stories over the course of complete stories, also referred to as \textit{emotional arcs}, using supervised machine learning, and, in particular, \acp{LLM}. While predicting such valence and arousal signals is common in the field of multimodal affect analysis~\cite{ringeval2019avec, stappen2021muse, Christ22-TM2}, it has not been applied to textual stories, yet. 

\section{Related Work}\label{sec:rw}

Various unsupervised, lexicon-based approaches to model emotional trajectories in narrative and literary texts have been proposed. With a lexicon-based method,~\citet{reagan2016emotional} identified six elementary sentiment-based emotional arcs such as \textit{rags-to-riches} in a corpus of about 1,300 books.
\citet{moreira-etal-2023-modeling} generate lexicon-based emotional arcs and demonstrate their usability in predicting the perceived literary quality of novels. %
Further Examples include the works of ~\citet{strapparava2004wordnet}, \citet{wilson2005recognizing},~\citet{kim2017prototypical} and~\cite{somasundaran2020emotion}. \revision{While these previous works use dictionaries to directly predict emotionality, we only utilize them to map existing annotations into the valence/arousal space.} 

Moreover, a range of datasets of narratives annotated for emotionality exists. 
In a corpus of $100$ crowdsourced short stories,~\cite{mori2019narratives} provided annotations both for character emotions as well as for emotions evoked in readers.
The \ac{DENS}~\cite{liu2019dens} contains about 10,000 passages from modern as well as classic stories, labeled with $10$ discrete emotions. In the authors' experiments, fine-tuning \textsc{BERT}~\cite{devlin2018bert} proved to be superior to more classic approaches such as \acp{RNN}.
The \ac{REMAN} dataset~\cite{kim2018feels} comprises 1,720 text segments from about $200$ books. These passages are labeled on a phrase level regarding, among others, emotion, the emotion experiencer, the emotion's cause, and its target.~\citet{kim2018feels} conducted experiments with biLSTMs and \acp{CRF} on \ac{REMAN}. 
The \ac{SEND}~\cite{ong2019modeling} is a multimodal dataset containing $193$ video clips of subjects narrating personal emotional events, annotated with valence values in a time-continuous manner.

The corpus of children's stories~\cite{alm2008affect} we are using for our experiments is originally labeled for eight discrete emotions (cf.~\Cref{sec:data}).~\citet{alm2005emotional} modeled emotional trajectories in a subset of this corpus, while in~\cite{alm2005emotions}, the authors conducted machine learning experiments with several handcrafted features such as sentence length and POS-Tags as well as Bag of Words. While the corpus has frequently served as a benchmark for textual emotion recognition, scholars have so far limited their experiments to subsets of this dataset, selected based on high agreement among the annotators or certain emotion labels. Examples of such studies include 
an algorithm combining vector representations and syntactic dependencies by~\citet{agrawal2012unsupervised}, 
the rule-based approach proposed by~\citet{udochukwu2015rule}, and a combination of \ac{CNN} and \ac{LSTM} introduced by~\citet{batbaatar2019semantic}. 
No existing work, however, aims at modeling the complete stories provided in the dataset. 

\section{Data}

\label{sec:data}
We choose the children's story dataset by~\citet{alm2008affect}, henceforth referred to as \textsc{Alm}, for our experiments. \revision{From the mentioned datasets, the \textsc{Alm} dataset is the only suitable one} as it is reasonably large, comprising about 15,000 sentences, and contains complete, yet brief stories, with the longest story consisting of $530$ sentences.
 Moreover, the data is labeled per sentence, allowing us to model emotional trajectories for stories. We extend the dataset by a third annotation, as described in~\Cref{ssec:annotation}, and modify the originally discrete annotation scheme by mapping it into the continuous valence/arousal space (cf.~\Cref{ssec:mapping}). 

 Originally, the dataset comprises $176$ stories from $3$ authors. More precisely, $80$ stories from the German \textit{Brothers Grimm}, $77$ stories by Danish author \textit{Hans-Christian Andersen}, and $19$ stories written by \textit{Beatrix Potter} are contained. Every sentence is annotated with the emotion experienced by the primary character (\textit{feeler}) in the respective sentence, and the overall mood of the sentence. For both label types, two annotators had to select one out of eight discrete emotion labels, namely \textit{anger}, \textit{disgust}, \textit{fear}, \textit{happiness}, \textit{negative surprise}, \textit{neutral}, \textit{positive surprise}, and \textit{sadness}. 
 For a detailed description of the original data, the reader is referred to~\cite{alm2005emotional, alm2008affect}. 
 We limit our experiments to predicting the mood per sentence, as it refers to the sentence as a whole instead of a particular subject.

\subsection{Additional Annotations}\label{ssec:annotation}
In addition to the existing annotations, we collect a third mood label for every sentence. This allows us to create a continuous-valued gold standard (cf.~\Cref{ssec:mapping}) via the agreement-based \ac{EWE}~\cite{grimm2005evaluation} fusion method, for which at least three different ratings are required. Compared to the original dataset, however, we utilize a reduced labeling scheme, eliminating both \textit{positive surprise} and \textit{negative surprise} from the set of emotions. We follow the reasoning of~\citet{susanto2020hourglass} and~\citet{ortony2022all}, who argue that \textit{surprise} can not be considered a basic emotion, as it is not \textit{valenced}, \ie of negative or positive polarity, in itself but can only be polarised in combination with other emotions. 

Krippendorff's alpha ($\alpha$) for all three annotators is $.385$, when calculated based on single sentences. Details on agreements are provided in~\Cref{appdx:agreements}.
Removal of $7$ low-agreement stories (cf.~\Cref{appdx:agreements}) leaves us with a final data set of $169$ stories. Key statistics of the data are summarized in~\Cref{tab:data}.

\begin{table}[]
    \centering
    \resizebox{1\columnwidth}{!}{
    \begin{tabular}{lrrrr}
    \toprule
         \multicolumn{1}{c}{} & \multicolumn{1}{c}{Overall} & \multicolumn{1}{c}{Grimm} & \multicolumn{1}{c}{HCA} & \multicolumn{1}{c}{Potter} \\ \midrule 

        \multicolumn{5}{l}{\textit{Size}} \\
        \# sentences & 14,884 & 5,236 & 7\,712 & 1,936 \\
         \# stories & 169 & 77 & 73 & 19 \\ 
\midrule
& \multicolumn{4}{l}{Emotion Distribution [\%]} \\
\midrule
         anger & 4.54 & 6.71 & 2.77 & 5.71 \\

         disgust & 2.35 & 1.78 & 2.83 & 1.98 \\

         fear & 7.21 & 11.48 & 3.77 & 9.38 \\

         happiness & 14.42 & 13.74 & 16.59 & 7.58\\

         negative surprise & 4.41 & 4.17 & 4.74 & 3.72 \\

         neutral & 56.19 & 49.88 & 57.86 & 66.56 \\

         positive surprise & 1.90 & 2.73 & 1.54 & 1.08 \\

         sadness & 8.99 & 9.51 & 9.89 & 3.97 \\
         \bottomrule
    \end{tabular}}
    \caption{Key statistics for the entire dataset and the subsets defined by the three different authors.}
    \label{tab:data}
\end{table}
The label distribution statistics listed in~\Cref{tab:data} indicate stylistic differences between the different authors. To give an example, 
 \textit{sadness} is rare in \textit{Potter's} stories ($3.97\,\%$ of all annotations) compared to the other two authors. Overall, \textit{neutral} is the most frequent label, while other classes, especially \textit{positive surprise} and \textit{disgust}, are underrepresented.

\begin{figure}[h!]
    \centering
    \includegraphics[width=\columnwidth]{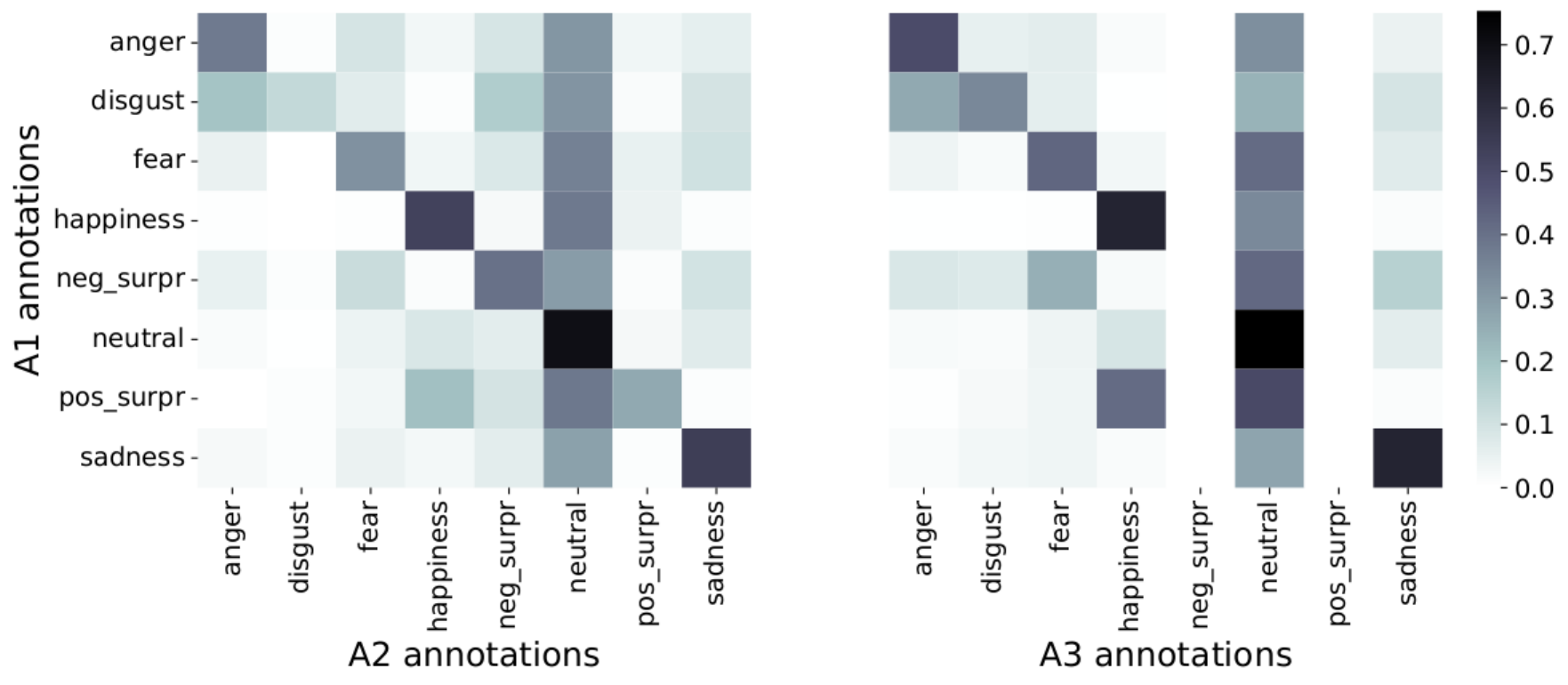}
    \caption{Confusion matrices comparing different annotators' (A1, A2, A3) labels for the whole dataset. Note that for annotator 3, \textit{positive} and \textit{negative surprise} were not available.}
    \label{fig:conf}
\end{figure}

\Cref{fig:conf} shows confusion matrices comparing the annotations of annotator 1 with the annotations of annotators 2 and 3. The decision of whether a sentence is emotional or \textit{neutral} is the most important source of disagreement in both annotator pairs. Furthermore,~\Cref{fig:conf} demonstrates that disagreement about the valence, \ie pleasantness, of a sentence's mood is rare. To give an example, in both depicted confusion matrices, sentences labeled with \textit{happiness} by annotator 1 are rarely labeled with a negative emotion (\textit{anger}, \textit{disgust}, \textit{fear}) by annotator 2 and 3, respectively.

\subsection{Label Mapping}\label{ssec:mapping}
Motivated by low to moderate Krippendorff agreements (cf.~\Cref{appdx:agreements}) and underrepresented classes in the discrete annotations (cf.~\Cref{tab:data}), we project all emotion labels into the more generic, continuous valence/arousal space. Proposed by~\cite{russell1980circumplex}, the valence/arousal model characterizes affective states among two continuous dimensions where valence corresponds to pleasantness, while arousal is the intensity or degree of agitation. As depicted in~\Cref{fig:conf}, the annotators often agree on the polarity of the emotion, \ie whether it is to be understood as positive or negative in terms of valence. Hence, it can be argued that disagreement between annotators is not always as grave as suggested by low $\alpha$ values, which do not take proximity between different emotions into account. To give an example, disagreement on whether a sentence's mood is \textit{happiness} or \textit{neutral} is certainly less severe than one annotator labeling the sentence \textit{sad}, while the other opts for \textit{happy}. Moreover, a projection into continuous space unifies the two different label spaces defined by the original and our additional annotations, respectively.
To implement the desired mapping, we take up an idea proposed by~\citet{park-etal-2021-dimensional}, who map discrete emotion categories to valence and arousal values by looking up the label (\eg \textit{anger}) in the NRC-VAD dictionary~\cite{mohammad2018obtaining} that assigns crowd-sourced valence and arousal values in the range $[0...1]$ to words. For instance, the label \textit{anger} is mapped to a valence value of $.167$ and an arousal value of $.865$. The full mapping and further explanations can be found in~\Cref{appdx:mapping}.
    
\begin{figure*}[h!]
    \centering
    \includegraphics[width=.85\linewidth, page=1]{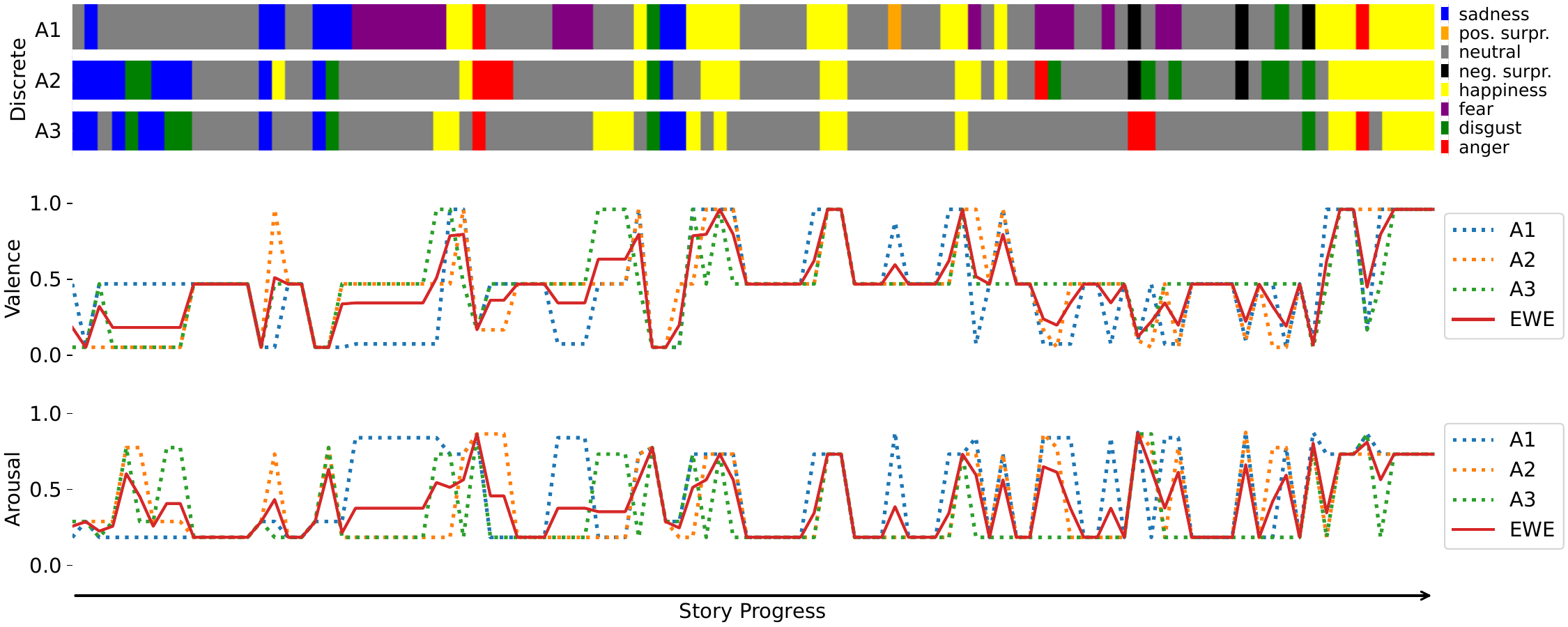}

    \caption{Exemplary mapping from the three annotators' (A1, A2, A3) discrete annotations (top) to their respective valence (middle) and arousal (bottom) signals and the gold standard signals created via EWE (solid red lines). The annotations are taken from the story \textit{Ashputtel} by the \textit{Grimm} brothers, consisting of 102 sentences.}\label{fig:signals}
\end{figure*}

After label mapping, we create a gold standard for every story by fusing the thus obtained signals over the course of a story for valence and arousal, respectively. We apply the \ac{EWE}~\cite{grimm2005evaluation} method which is well-established for the problem of computing valence and arousal gold standards from continuous signals (\eg\cite{ringeval2019avec, stappen2021muse, christ2022multimodal}).~\Cref{fig:signals} presents an example for this process, presenting both the discrete labels and the valence and arousal signals constructed from them for a specific story.
     
\subsection{Split}\label{ssec:splits}
We split the data on the level of stories.
Three partitions for training, development, and test are created, with $118$, $25$, and $26$ stories, respectively.
In doing so, we make sure to include comparable portions of stories and sentences by each author in all three partitions.
A detailed breakdown is provided in~\Cref{appdx:split}.

\section{Experimental Setup}
\label{sec:exp-setup}

We fine-tune (cf.~\Cref{ssec:finetuning}) the $304$M parameter \textit{large} version of \deberta~\cite{he2023debertav}, additionally utilizing a weakly supervised learning approach (\Cref{ssec:weakly}). 
Further details regarding the computational resources can be found in~\Cref{appdx:computation}.

\subsection{Finetuning}\label{ssec:finetuning}
\begin{figure}[h!]
    \centering
    \includegraphics[width=.95\columnwidth, trim=50 170 505 35, clip]{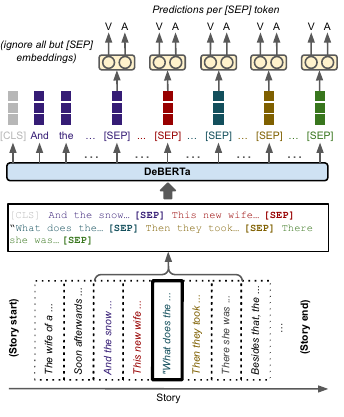}
    \caption{Example for the finetuning approach with context size $\mathcal{C}=2$. Valence (\textit{V}) and arousal (\textit{A}) predictions are obtained for all sentences at once.}
    \label{fig:finetuning}
\end{figure}

Since the context of a sentence in a story is relevant to the mood it conveys, we seek to leverage multiple sentences at once in the fine-tuning process. Specifically, we create training examples as follows. We denote a story as a sequence of sentences $s_1\ldots s_n$. For a sentence $s_i$ and a context window size $\mathcal{C}$, we also consider the up to $\mathcal{C}$ sentences preceding ($s_{i-\mathcal{C}}\ldots s_{i-1}$) and the up to $\mathcal{C}$ sentences following ($s_{i+1}\ldots s_{i+\mathcal{C}}$) $s_i$. We construct an input string from the sentences $s_{i-\mathcal{C}}\ldots s_{i+\mathcal{C}}$ by concatenating them, separated via the special \verb|[SEP]| token. The $i$-th \verb|[SEP]| token in this sequence is intended to represent the $i$th sentence. We add a token-wise feed-forward layer on top of the last layer's token representations. It projects each $1024$-dimensional embedding to $2$ dimensions and is followed by Sigmoid activation for both of them, corresponding to a prediction for valence and arousal, respectively. %
As the loss function, we sum up the \acp{MSE} of valence and arousal predictions for each \verb|[SEP]| token. We optimize $\mathcal{C}$, for $\mathcal{C}\in\{1, 2, 4, 8\}$. If the length of an input exceeds the model's capacity, we decrease $\mathcal{C}$ for this specific input. \Cref{fig:finetuning} provides an example of an input sequence. %

We train the models for at most $10$ epochs but abort the training process early if no improvement on the development set is achieved for $2$ consecutive epochs. The evaluation metric is the mean of the \ac{CCC}~\cite{lawrence1989concordance} values achieved for arousal and valence, computed over the whole dataset, respectively.~\Cref{eq:ccc} gives the formula for \ac{CCC} between two signals $Y$ and $\hat{Y}$ of equal length.

\begin{equation}\label{eq:ccc}
    \textit{CCC}(Y, \hat{Y}) = \frac{2~\textit{Cov}(Y, \hat{Y})}{\textit{Var}(Y)+\textit{Var}(\hat{Y})+(\overline{Y}-\overline{\hat{Y}})^2}
\end{equation}

\ac{CCC} is a well-established correlation measure to assess agreement between (pseudo-)time continuous annotations and predictions, particularly common in Affective Computing Tasks, \eg \cite{ringeval2018avec, schoneveld2021leveraging, christ2023muse}. \revision{It can be thought of as a bias-corrected modification of Pearson’s correlation. Different from Pearson's correlation, it is sensitive to location and scale shifts, \ie it measures not only correlation but also takes into account absolute errors. Same as for the Pearson correlation, the
chance level is $0$, and two identical signals would have a CCC value of $1$.} 
AdamW~\cite{loshchilov2018decoupled} is chosen as the optimization method. Following a preliminary hyperparameter search, the learning rate is set to $5\times10^{-6}$. We do not optimise any hyperparameter besides the learning rate. Every experiment is repeated with five fixed seeds. In every experiment, we initialize the model with the checkpoint provided by the \deberta authors\footnote{\href{https://huggingface.co/microsoft/deberta-v3-large}{https://https://huggingface.co/microsoft/deberta-v3-large}}.

\subsection{Weakly Supervised Learning}\label{ssec:weakly}
The Alm dataset comprises $169$ stories by only three different authors, making our models prone to overfitting. Thus, we seek to augment our data set with in-domain texts written by other authors, thereby covering more topics and also spanning more cultures.
We collect $45$ books containing in total $801$ different stories from Project Gutenberg, more specifically the \textit{Children's Myths, Fairy Tales, etc.} category\footnote{\href{https://www.gutenberg.org/ebooks/bookshelf/216}{https://www.gutenberg.org/ebooks/bookshelf/216}}. 
These stories comprise fairytales, myths, and other tales from different geographic regions, including Japan, Ireland, and India. This newly collected unlabeled data set, henceforth referred to as \emph{Gutenberg Corpus} or \textsc{Gb}, amounts to $101529$ sentences. A more detailed description of \textsc{Gb} is given in~\Cref{appdx:gb}. 

\begin{figure}[h!]
    \centering
    \includegraphics[width=.97\columnwidth, angle=0, trim=0 100 415 37, clip]{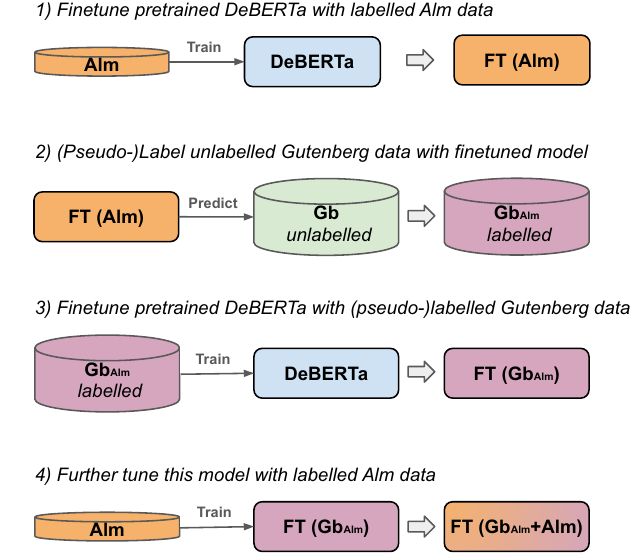}
    \caption{Illustration of our training steps and corpora. \textit{FT} is short for \textit{finetuned}.}
    \label{fig:framework}
\end{figure}

\Cref{fig:framework} illustrates our overall finetuning approach. Given 1) a DeBERTa model finetuned on the labeled dataset (cf.~\ref{ssec:finetuning}), we 2) utilize its predictions on \textsc{Gb} as pseudo-labels, yielding a labeled dataset \textsc{Gb}{\tiny \textsc{Alm}}.
Subsequently, 3) another pretrained DeBERTa model is finetuned on \textsc{Gb}{\tiny\textsc{Alm}} only. This training process is limited to $1$ epoch and employs a learning rate of $5\times10^{-1}$. Lastly, 4) this model is further trained on \textsc{Alm}. Here, we utilize the same hyperparameters as for training $M$, but we find a smaller learning rate of $10^{-7}$ to be beneficial.

\section{Results}

\begin{table*}[h!]
\centering 
\resizebox{.80\linewidth}{!}{
\begin{tabular}{l|rrrr|rrrr|rrrr}
\toprule
 & \multicolumn{4}{c|}{\textbf{FT \textsc{Alm} [CCC$\uparrow$]}} & \multicolumn{4}{c|}{\textbf{FT \textsc{Gb}{\tiny\textsc{Alm}} [CCC$\uparrow$]}} & \multicolumn{4}{c}{\textbf{FT \textsc{Gb}{\tiny Alm} +  \textsc{Alm} [CCC$\uparrow$]}}\\
 
 & \multicolumn{2}{c}{Valence} & \multicolumn{2}{c|}{Arousal} & \multicolumn{2}{c}{Valence} & \multicolumn{2}{c|}{Arousal} & \multicolumn{2}{c}{Valence} & \multicolumn{2}{c}{Arousal}  \\

 $\mathcal{C}$ & \multicolumn{1}{c}{dev} & \multicolumn{1}{c}{test} & \multicolumn{1}{c}{dev} & \multicolumn{1}{c|}{test} & \multicolumn{1}{c}{dev} & \multicolumn{1}{c}{test} & \multicolumn{1}{c}{dev} & \multicolumn{1}{c|}{test} & \multicolumn{1}{c}{dev} & \multicolumn{1}{c}{test} & \multicolumn{1}{c}{dev} & \multicolumn{1}{c}{test} \\ \midrule

0 & .6798 & .6972 & .5576 & .5904 & 
.6967 & .7134 & .5744 & .6109 & 
\uline{.7007} & \uline{.7184} & \uline{.5793} & \uline{.6168}\\ 

1 & .7522 & .7538 & .6206 & .6593 & 
.7658 & .7782 & .6365 & .6782 & 
\uline{.7712} & \uline{.7842} & \uline{.6436} & \uline{.6873}\\ 

2 & .7746 & .7812 & .6404 & .6752 & 
.7857 & .7986 & .6541 & .6942 & 
\uline{.7905} & \uline{.8024} & \uline{.6626} & \uline{.7018}\\ 

4 & .7924 & .7881 & .6550 & .6762 & 
.7997 & .8056 & .6665 & .6989 & 
\uline{.8059} & \uline{.8138} & \uline{.6742} & \uline{.7098}\\ 

8 & \textbf{.7983} & \textbf{.7988} & \textbf{.6613} & \textbf{.6798} & 
\textbf{.8098} & \textbf{.8141} & \textbf{.6736} & \textbf{.7027} & 
\uline{\textbf{.8168}} & \uline{\textbf{.8221}} & \uline{\textbf{.6809}} & \uline{\textbf{.7125}}\\ 

\bottomrule

 \end{tabular}
 }
 \caption{Results for fine-tuning (\textit{FT}) with different context sizes $\mathcal{C}$. See \Cref{fig:framework} for an illustration of the three different corpora. The results are averaged over 5 fixed seeds. Standard deviations are negligible and thus omitted. Overall, the best results on the development set per prediction target and partition are boldfaced, and the best results for each context size are underlined.} \label{tab:main_results}\end{table*}

\Cref{tab:main_results} presents the results of our experiments with different $\mathcal{C}$ values. We report the mean \ac{CCC} results when tuning a) only on \textsc{Alm} (FT \textsc{Alm}, step 1 in~\Cref{fig:framework}), b) only on \textsc{Gb}{\tiny\textsc{Alm}} (FT \textsc{Gb}{\tiny\textsc{Alm}}, step 3 in~\Cref{fig:framework}) and c) additionally on \textsc{Alm} (FT \textsc{Gb}{\tiny Alm} +  \textsc{Alm}, step 4 in~\Cref{fig:framework}). There is a clear trend for both arousal and valence to increase with larger $\mathcal{C}$s. The models trained with a context size of $8$ account for the best valence and arousal results in every set of experiments, \eg \ac{CCC} values of $.8168$ and $.6809$ for arousal and valence, respectively, on the development set when trained on both corpora. These are also the best results encountered overall. In contrast, the models with $\mathcal{C}=0$ always perform worst, yielding \eg only \ac{CCC} values of $.6798$ (valence) and $.5576$ (arousal) on the development set in the \textsc{Alm}-only configuration. This supports the assumption that the context of a sentence is oftentimes key to correctly assessing its mood. The gap between valence and arousal \ac{CCC} values is in line with previous studies showing that text-based classifiers are typically better suited for valence prediction than for arousal prediction~\cite{kossaifi2019sewa, wagner2023dawn}. Further, our results demonstrate the benefits of the weakly supervised approach. Training on \textsc{Gb}{\tiny{\textsc{Alm}}} always improves upon training on \textsc{Alm} only, especially for smaller context sizes $\mathcal{C}$. To give an example, both the valence and arousal \ac{CCC} values on the development set increase by more than $1.5$ for $\mathcal{C}=0$ on the development set. Further tuning on \textsc{Alm} afterward leads to additional performance gains for both prediction targets and all context sizes. However, the increase never exceeds $1$ percentage point in comparison to training on \textsc{Gb}{\tiny{\textsc{Alm}}}.

\subsection{Author-Wise Results}

\begin{table}[h!]
 \resizebox{.95\columnwidth}{!}{
\centering
\begin{tabular}{lrrrrrr}
\toprule 
\multicolumn{1}{l}{[CCC$\uparrow$]} & \multicolumn{5}{c}{\textsc{Auth}}\\ & \multicolumn{2}{c}{Grimm} &  \multicolumn{2}{c}{HCA} & \multicolumn{2}{c}{Potter} \\ 
\midrule
Finetuning (FT) & \multicolumn{1}{c}{Valence} &  \multicolumn{1}{c}{Arousal} & \multicolumn{1}{c}{Valence} &  \multicolumn{1}{c}{Arousal} & \multicolumn{1}{c}{Valence} &  \multicolumn{1}{c}{Arousal} \\
\midrule

\textsc{Alm} & .7943 & .6804 & .7942 & .6881 & .7606 & .6442 \\
\textsc{Gb}{\small \textsc{Alm}} & .8131 & .7104 & .8247 & .7253 & .7672 & .6761 \\
\hdashline
\textsc{Alm}$\setminus\{$\textsc{Auth}$\}$  & .7597 & .6505 & .7633 & .6563 & .7304 & .5886 \\
\textsc{Gb}{\small \textsc{Alm}$\setminus\{$\textsc{Auth}$\}$} & .7799 & .6734 & .7837 & .6774 & .7386 & .6344 \\

\bottomrule
\end{tabular}}
\caption{Author-wise experiment results on the respective test sets. The results are averaged and standard deviations (all $<.01$) are omitted. Cf.~\Cref{fig:framework} for illustration of the methods.}\label{tab:auth_analysis}
\end{table}

Since \textsc{Alm} comprises stories of three different authors, we investigate the relevance of an author's individual style for learning to predict their stories.
For each author (\textsc{Auth}), we create a dataset \textsc{Alm}$\setminus\{\textsc{Auth}\}$ by removing \textsc{Auth}'s stories from the training and development partitions and keeping only \textsc{Auth}'s stories as test data. We then repeat steps 1-3 in ~\Cref{fig:framework}, using \textsc{Alm}$\setminus\{\textsc{Auth}\}$ instead of the full \textsc{Alm} dataset. Only the best configuration, \ie $\mathcal{C}=8$ is considered here.
The results of these experiments, alongside the corresponding author-wise results, when employing the full \textsc{Alm} dataset, are given in~\Cref{tab:auth_analysis}.
Performance, in general, differs by author, \eg both valence and arousal \ac{CCC} for \textit{Potter} are lower than for the other two authors when training on the full \textsc{Alm} dataset.
Furthermore, test set performance for every author drops when removing the author from the training and development data. The clearest example is \textit{Potter}'s arousal \ac{CCC} value of $.5886$ when training on \textsc{Alm}$\setminus\{\textit{Potter}\}$ compared to $.6442$ when training on the full \textsc{Alm} data.
The weakly supervised learning step, implying exposure to a wider range of styles, proves to be beneficial for every author, regardless of the dataset. Nevertheless, for every author \textsc{Auth} the performance of the weakly supervised approach on \textsc{Alm}$\setminus\{\textsc{Auth}\}$ never reaches the performance for fine-tuning on \textsc{Alm} alone. In conclusion, it is crucial to include targeted authors in training data in order to capture their individual styles.

\subsection{Further Statistics}\label{ssec:analysis_author}

In the remainder of the paper, we limit our analysis to the best-performing seed for $\mathcal{C}=8$ and the full training pipeline (cf.~\Cref{fig:framework}).

The \ac{CCC} values given in~\Cref{tab:main_results} are calculated over the entire dataset, \ie a concatenation of all stories per partition.~\Cref{tab:story_wise}, in contrast, lists \emph{story-wise} \ac{CCC} results for the predictions of our best model.

\begin{table}[h!]\centering
 \resizebox{.8\columnwidth}{!}{

\begin{tabular}{lrr}
\toprule 
\multicolumn{1}{l}{Partition} &  \multicolumn{1}{r}{Valence [CCC$\uparrow$]} & \multicolumn{1}{r}{Arousal [CCC$\uparrow$]} \\ 
\midrule

Dev  & .7729 ($\pm$.1207) & .6892 ($\pm$.0812) \\
Test & .7685 ($\pm$.0946) & .6352 ($\pm$.1679) \\
\bottomrule
\end{tabular}}
\caption{Story-wise CCC results over all stories in the development and test set as predicted by the best model.}\label{tab:story_wise}
\end{table}

It shows that results are highly story-dependent. To give an example, the arousal \ac{CCC} values for the test partition display a standard deviation of $.1679$ over the $25$ stories in this partition. 
 
  We find that the model's performance for arousal and valence per story correlates: we obtain a Pearson's correlation of $.3499$ (statistically significant with $p<.02$) between our best model's valence \ac{CCC} values per story and the respective \ac{CCC} arousal values. Hence, there exist stories whose emotional trajectories are difficult (or easy) to predict for the model in general, regardless of the two different emotional dimensions.  
  
 This can partly be explained by the correlation between model performance and human agreement per story. There is a Pearson's correlation of $.3659$ between all story-wise human \ac{CCC} agreements and the \ac{CCC} values achieved by the best model on the corresponding stories. Analogously, for arousal, this correlation is $.4095$. Both correlations are statistically significant with $p<.02$. It can be concluded that the model particularly struggles to learn stories that also pose a challenge to humans.

 Another analysis reveals that the model's performance also tends to vary for \emph{different parts of the same story}. We divide every story into 5 parts of equal size. This way, we evaluate the performance of our best models on 5 different subsets of the data corresponding to positions in the story. Roughly, the first part can be expected to correspond to the beginning of the story, while the last part comprises its end.~\Cref{tab:pos_analysis} displays the results of this evaluation. Both valence and arousal results are, on average, better at the very beginning ($.8260$ valence, $.7304$ arousal \ac{CCC}) and the very end ($.8576$ valence, $.7306$ arousal \ac{CCC}) of the stories than during their middle parts. We hypothesize that this is due to many stories' beginnings and endings being drawn from a limited set of archetypical situations. Hence, the model may easily learn the emotional connotations of such common \revision{events} from the large corpus. To give a few examples, fairytales in particular often start, \eg with the death or absence of a parent, the hero leaving home, an act of villainy against the hero, or a combination thereof. Endings often involve reunion, marriage, and the villain receiving punishment~\cite{propp1968morphology}.

 As a measure of model performance on the sentence level, we compute the best model's absolute prediction errors on the development and test set. \Cref{tab:maes} presents the results.
 
 \begin{table}[h!]\centering
 \resizebox{.95\columnwidth}{!}{

\begin{tabular}{lrrrrrr}
\toprule 
 & \multicolumn{1}{l}{mean} & \multicolumn{1}{l}{std} & \multicolumn{1}{l}{median} & \multicolumn{1}{l}{perc. $90$} & \multicolumn{1}{l}{perc. $95$} \\
\midrule
Valence & .1001 & .0971 & .0721 &  .2302 & .2940 \\ 
Arousal & .1205 & .1144 & .0838 &  .2878 & .3626 \\ 

\bottomrule
\end{tabular}}
\caption{Absolute error statistics for the development and test data predictions (combined) of the best model.}\label{tab:maes}
\end{table}
 
It is evident from the median values that more than half of arousal and valence predictions miss the gold standard by less than $.1$. From the percentiles, it can be concluded that errors larger than $.3$ occur in less than $5\,\%$ of sentences for valence and in less than $10\,\%$ of sentences for arousal.

\begin{table}[h!] \centering
 \resizebox{.95\columnwidth}{!}{
\begin{tabular}{lrrrrr}
\toprule 
\multicolumn{1}{c}{[CCC$\uparrow$]} & \multicolumn{5}{c}{Story Part} \\
 &  \multicolumn{1}{l}{$1$/$5$} & \multicolumn{1}{l}{$2$/$5$} & \multicolumn{1}{l}{$3$/$5$} & \multicolumn{1}{l}{$4$/$5$} & \multicolumn{1}{l}{$5$/$5$} \\ 
\midrule
Valence & .8260 & .8044 & .8091 & .7785 & .8576 \\ 
Arousal & .7304 & .6543 & .6814 & .6618 & .7306\\

\bottomrule
\end{tabular}}
\caption{Mean \ac{CCC} values across 5 seeds for the best configuration on different story parts. We omit the low standard deviations(all $<1$), omitted. Results were computed over the unification of test and dev.}\label{tab:pos_analysis}
\end{table}

\subsection{Qualitative Analysis}\label{ssec:analysis_qualitative}

To gain qualitative insights into the model's limitations, we manually analyze around $200$ text spans for which high absolute errors in terms of valence or arousal prediction are observed. First, we find that the model seems to learn emotional connotations of events, but is prone to ignore the roles of the protagonists involved in them. \Cref{tab:example_house} provides an example of this phenomenon. In this text passage, a typically positive event, namely being granted a wish, is salient. The model assigns relatively high valence values. However, the actual mood in these sentences is rather negative, as they describe the implicitly jealous reaction of a negative character to this situation.

\begin{table}[h!]\centering
 \resizebox{.98\linewidth}{!}{

\begin{tabular}{lrr}
\toprule
\midrule 

\multicolumn{3}{l}{\textbf{Story}: \textit{87\_the\_poor\_man\_and\_the\_rich\_man} (Grimms)} \\

\multicolumn{3}{l}{\textbf{Context}: A poor man receives a new house as a reward from God. His } \\
\multicolumn{3}{l}{neighbor (rich man) sees the new house and gets jealous.} \\

\midrule
\midrule

\multicolumn{1}{l}{\textbf{Sentence}} &  \multicolumn{1}{r}{\textbf{V pred}} & \multicolumn{1}{r}{\textbf{V GS}} \\ 
\midrule

\multirow{4}{250pt}{
The sun was high when the rich man got up and leaned out of his window and saw, on the opposite side of the way, a new clean-looking house with red tiles and bright windows where the old hut used to be.

}  & .8044 & .4690 \\ 
& & \\
& & \\
& & \\
\midrule

\multirow{2}{250pt}{He was very much astonished, and called his wife and said to her, "Tell me, what can have happened?} 
& .5657 & .2220 \\
& & \\
\midrule

\multirow{2}{250pt}{Last night there was a miserable little hut standing there, and to-day there is a beautiful new house.} & .7744 & .2220 \\
& & \\

\midrule
\bottomrule
\end{tabular}}
\caption{Passage from \textit{87\_the\_poor\_...} (Grimms) with valence (\textit{V}) predictions (\textit{pred}) and gold standard (\textit{GS}).}\label{tab:example_house}
\end{table}

Probably closely related to these observations, we figure that our model sometimes struggles to accurately assess situations, because it disregards the general sentiment of the respective story. To give an example,~\Cref{tab:example_grandmot} lists a passage from Andersen's story \textit{grandmot}. This story displays a rather positive sentiment overall, as it is presented as a loving memory of a deceased grandmother. For the passages describing her peaceful death, our model underestimates the valence gold standard by a large margin, probably due to the typically sad topic of death.

\begin{table}[h!]\centering
 \resizebox{.98\linewidth}{!}{

\begin{tabular}{lrr}
\toprule
\midrule 

\multicolumn{3}{l}{\textbf{Story}: \textit{grandmot} (Andersen)} \\

\multicolumn{3}{l}{\textbf{Context}: The story is about a beloved grandmother and her peaceful death}\\

\midrule
\midrule

\multicolumn{1}{l}{\textbf{Sentence}} &  \multicolumn{1}{r}{\textbf{V pred}} & \multicolumn{1}{r}{\textbf{V GS}} \\ 
\midrule

\multirow{1}{250pt}{She smiled once more, and then people said she was dead.}   & .3053 & .6470\\
\midrule

\multirow{4}{250pt}{She was laid in a black coffin, looking mild and beautiful
[...] though her eyes were closed;
but every wrinkle had vanished, her hair looked white and silvery,
and around her mouth lingered a sweet smile.}  & .2902& .7910 \\
& & \\
& & \\
& & \\
\midrule

\multirow{2}{250pt}{We did not feel at all afraid to look at the corpse of her who had been such a dear, good grandmother.}  & .3197 & .4700\\
& & \\

\midrule
\bottomrule
\end{tabular}}
\caption{Passage from \textit{grandmot} (Andersen) with valence (\textit{V}) predictions (\textit{pred}) and gold standard (\textit{GS}).}\label{tab:example_grandmot}
\end{table}
    
   Stories within stories pose another facet the model faces difficulties with. Frequently, protagonists tell stories or recall memories. Narrated stories or memories typically contain emotionally significant events, but they are not directly experienced and thus are not always heavily influencing the mood of the actual story.~\Cref{tab:example_rat} presents an example, where a cat tells another one about a fearful incident. The corresponding gold standard arousal values are moderate, arguably as the incident is over and has not harmed the protagonist. The model nevertheless predicts high arousal.
\begin{table}[h!]\centering
 \resizebox{.98\linewidth}{!}{
\begin{tabular}{lrr}
\toprule
\midrule 

\multicolumn{3}{l}{\textbf{Story}: \textit{the\_roly-poly\_pudding} (Potter)} \\

\multicolumn{3}{l}{\textbf{Context}: A cat tells another cat (\textit{Ribby}) how he had encountered a rat.}\\

\midrule
\midrule

\multicolumn{1}{l}{\textbf{Sentence}} &  \multicolumn{1}{r}{\textbf{A pred}} & \multicolumn{1}{r}{\textbf{A GS}} \\ 
\midrule

\multirow{2}{250pt}{I caught seven young ones [rats] [...], and we had them for dinner last Saturday.}  & .6566 & .1840 \\
 & & \\
\midrule

\multirow{2}{250pt}{And once I saw the old father rat --an enormous old rat -- Cousin Ribby.} 
& .5070 & .3940 \\
 & & \\
\midrule

\multirow{2}{250pt}{I was just going to jump upon him, when he showed his yellow teeth at me and whisked down the hole.} & .7283 & .1840 \\
 & & \\

\midrule
\bottomrule
\end{tabular}}
\caption{Passage from \textit{the\_roly\_poly...} (Potter) with arousal (\textit{A}) predictions (\textit{pred}) and gold standard (\textit{GS}).}\label{tab:example_rat}
\end{table}

To summarise, the model tends to miss out on a holistic understanding of stories such as the roles of different protagonists, nested stories, and a story's overall tone. This can partially be attributed to inputs not consisting of complete stories, cf.~\Cref{ssec:finetuning}.
Further examples for all aspects discussed above can be found in~\Cref{appdx:analysis}.

\section{Discussion}
We demonstrate the efficacy of our approach to model emotional trajectories via \acp{LLM}, achieving \ac{CCC} values of $.8221$ and $.6809$ for valence and arousal on the test set, respectively. We find that considering a sentence's context is crucial for predicting its emotionality. Furthermore, our analysis reveals the author-dependence of these results, which, in addition, vary from story to story. Even within a story, certain parts (namely, beginning and ending) are often easier to predict than others. Further analysis of our models' predictions uncovers additional challenges, such as assigning the correct role to protagonists and understanding the overall tone of a story. All these aspects combined shed light on the complexity of the task at hand. Keeping this in mind, our methodology can be understood as a first benchmark for predicting emotional trajectories in a supervised manner.

\section{Conclusion}
We proposed a valence/arousal-based gold standard for the Alm dataset~\cite{alm2008affect}. Moreover, we provide first results for the prediction of these signals via finetuning DeBERTa combined with a weakly supervised learning step. We obtain promising results, but, at the same time, demonstrate the limits of this methodology in our analysis. Future work may include attempts at a more holistic story understanding, involving \eg the roles of protagonists. Besides, our analysis of the results by author suggests that personalization methods (\eg \cite{kathan2022personalised}) may improve the results. Further, the potential of even larger \acp{LLM} such as LLaMA~\cite{touvron2023llama} or \mbox{(Chat-)GPT}~\cite{achiam2023gpt} remains to be explored for this task. Such models may even assist in refining the rather simplistic mapping method we utilized for the creation of the gold standard, as they have been shown to come with inherent emotional understanding capabilities~\cite{broekens2023fine, Tak2023IsGA}.
Code, data, and model weights are released to the public\footnote{\href{https://github.com/lc0197/emotional_trajectories_stories}{https://github.com/lc0197/emotional\_trajectories\_stories}}.

\section{Limitations}
Our work 
comes with several constraints. The simple mapping from discrete emotions into the dimensional valence/arousal space (cf.~\Cref{tab:mapping}) may be too coarse to capture some texts' emotional connotations. 
When analyzing the best model's predictions, we encounter texts where such shortcomings of our label mapping approach (cf.~\Cref{tab:mapping}) surface. For instance, all instances of \textit{disgust} are mapped to high arousal, leaving no room for less frequent low-arousal variants of \textit{disgust} as can be found in passages like the one given in~\Cref{tab:example_disgust}. Here, the sentences' mood was predominantly labeled as \textit{disgust} and thus set to high gold standard arousal values. The considerably lower arousal predictions by our model, however, are arguably more appropriate than the gold standard here, as the described situation is rather characterized by distanced arrogance than by actual \textit{disgust}.
 
\begin{table}[h!]\centering
 \resizebox{.95\linewidth}{!}{

\begin{tabular}{lrr}
\toprule
\midrule 

\multicolumn{1}{l}{\textbf{Story}: \textit{good\_for} (Andersen)} \\

\multicolumn{3}{l}{\textbf{Context}: An arrogant man sees a drunk poor woman from his window.
}\\

\midrule
\midrule

\multicolumn{1}{l}{\textbf{Sentence}} &  \multicolumn{1}{r}{\textbf{A pred}} & \multicolumn{1}{r}{\textbf{A GS}} \\ 
\midrule

\multirow{2}{250pt}{``Oh, it is the laundress,'' said he; ``she has had a little drop too much.}  & .3335 & .8050 \\ 
& & \\
\midrule

\multirow{1}{250pt}{She is good for nothing.} 
& .5281 & .6340 \\
\midrule

\multirow{1}{250pt}{It is a sad thing for her pretty little son.} & .3242 & .6700 \\

\midrule
\bottomrule
\end{tabular}}
\caption{Passage from \textit{good\_for} (Andersen) with arousal predictions (\textit{pred}) and gold standard (\textit{GS}).}\label{tab:example_disgust}
\end{table}

Besides, our approach to weakly supervised learning is obviously limited to high-resource languages. \textit{Story} is a broad term applicable to all texts in both \textsc{Alm} and the crawled \textsc{Gb} data. The included stories could be distinguished in a more fine-grained manner, \eg the data contains fairytales, myths, fables, and other types of stories. Such distinctions may have methodologically relevant implications we do not consider in our experiments. We also show that emotional arcs are highly author-dependent (cf.~\Cref{ssec:analysis_author}), implying that future datasets should seek to comprise a wider range of authors and writing styles. In particular, our results may not generalize well to authors of backgrounds that are not represented in the data used. Lastly, we analyse our method's limitations in~\Cref{ssec:analysis_qualitative}, without claim of completeness.

\section*{Acknowledgements}
Shahin Amiriparian, Manuel Milling, and 
Bj\"orn W. Schuller are also with the Munich Center for Machine Learning (MCML). Additionally, Bj\"orn W. Schuller is with the Munich Data Science Institute (MDSI) and the Konrad Zuse School of Excellence in Reliable AI (relAI), all in Munich, Germany and acknowledges their support.

\bibliography{custom}
\newpage
\appendix

\section{Annotation Details}

\begin{figure*}[h!]
    \centering
    {
    \includegraphics[width=.9\linewidth]{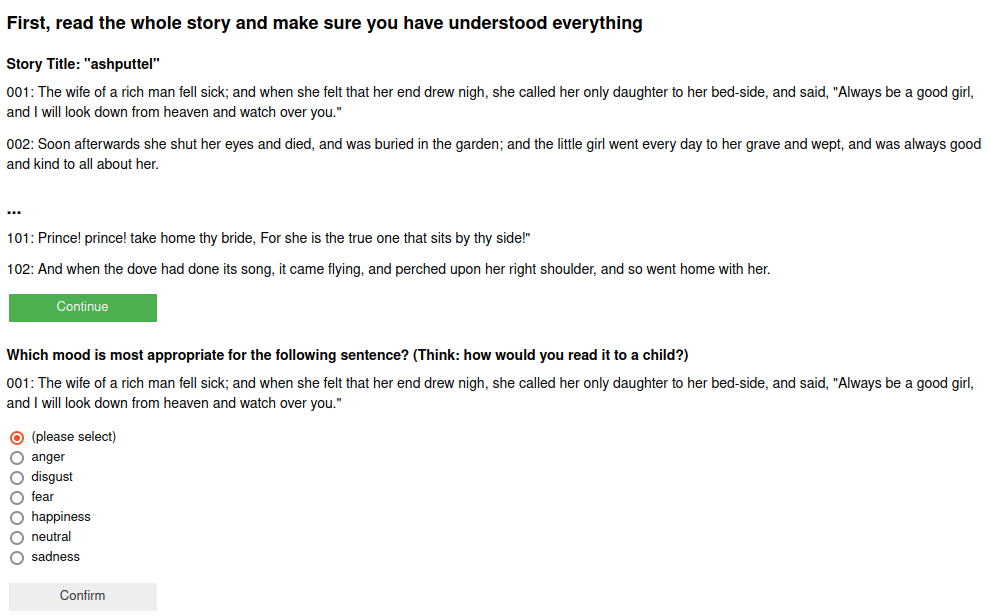}}
    \caption{Screenshot of the annotation tool. First, the whole story must be read. Upon confirmation (``Continue''), annotation of the individual sentences follows.}
    \label{fig:tool}  
\end{figure*}

Our additional annotations (cf.~\Cref{ssec:annotation}) are carried out by a 24-year-old male PhD student with a solid background in Affective Computing concepts, in particular, different emotion models. Hence, A3 is the same person for all stories, while this is not the case for A1 and A2 (cf.~\cite{alm2005emotional, alm2008affect}). \Cref{fig:tool} shows a screenshot of the annotation tool.

\section{Agreement Statistics}
\label{appdx:agreements}
Krippendorff's alpha ($\alpha$) for all three annotators is $.385$, when calculated based on single sentences and ignoring the different label schemes. This is possible, as annotator 3's label scheme is a subset of the labels available to annotators 1 and 2. The mean $\alpha$ per story is $\mu_\alpha=.341$, with a standard deviation of $\sigma_\alpha=.126$, indicating that the level of agreement is highly dependent on the story. We remove stories whose $\alpha$ is smaller than $\mu_\alpha - 2\sigma_\alpha$. A detailed listing of $\alpha$ values for the remaining data on both the sentence and the story level is provided in~\Cref{tab:agreement}.

\begin{table}[h!]
     \centering
     \resizebox{1\columnwidth}{!}{
\begin{tabular}{llrrrr}

\toprule 
Annotators & Level & \multicolumn{1}{r}{Overall} & \multicolumn{1}{r}{Grimms} & \multicolumn{1}{r}{HCA} & \multicolumn{1}{r}{Potter} \\ 
\midrule 

\multirow{2}{*}{A1,A2} & sent. & .356 & .272 & .411 & .333 \\ 
 & story & .297 ($\pm$.174) & .245 ($\pm$.196) &  .350 ($\pm$.149) & .307 ($\pm$.073) \\
 \midrule 

 \multirow{2}{*}{A1,A3} & sent. & .420 & .370 & .447 & .433 \\ 
 & story & .376 ($\pm$.184) & .346 ($\pm$.212) &  .395 ($\pm$.158) & .428 ($\pm$.126) \\
 \midrule 

 \multirow{2}{*}{A2,A3} & sent. & .383 & .331 & .408 & .391 \\ 
 & story & .338 ($\pm$.176) & .296 ($\pm$.172) &  .376 ($\pm$.178) & .3614 ($\pm$.139) \\
 \midrule 

 \multirow{2}{*}{A1,A2,A3} & sent. & .387 & .325 & .422 & .390 \\ 
 & story & .343 ($\pm$.126) & .301 ($\pm$.133) &  .380 ($\pm$.118) & .370 ($\pm$.062) \\
 \midrule 

\end{tabular}} \caption{$\alpha$ values for all possible combinations of annotators. The values are given for the whole dataset (\textit{Overall}) and the individual authors (\textit{Grimms, HCA, Potter}). The \textit{sent.} rows report the alphas on the basis of sentence annotations, in \textit{story} rows, the means, as well as standard deviations of alpha values per story, can be found.} \label{tab:agreement} 
     \end{table}
     
\Cref{tab:agreement} illustrates that agreement is also author-dependent, \eg for all combinations of annotators, the sentence-wise agreement for the \textit{Grimm} brothers is lower than for both other authors.

\section{Label Mapping Details}\label{appdx:mapping}

~\Cref{tab:mapping} lists the mapping for all discrete emotion labels as obtained from the NRC-VAD dictionary~\cite{mohammad2018obtaining}. However, the dictionary does not contain entries for \textit{positive surprise} and \textit{negative surprise}. For \textit{positive surprise}, we take the valence and arousal values of \textit{surprise} (both $.875$). The valence value for \textit{negative surprise} is set to the mean valence value of the negative emotions \textit{anger}, \textit{disgust},  and \textit{fear} ($.097$), while the arousal value is the same as for \textit{positive surprise} ($.875$).
    \begin{table}[h!]\centering
 \resizebox{.7\columnwidth}{!}{
\begin{tabular}{lrr}
\toprule 
\multicolumn{1}{l}{Label} &  \multicolumn{1}{r}{Valence} & \multicolumn{1}{r}{Arousal} \\ 

\midrule

Anger & .167 & .865 \\
Disgust & .052 & .775 \\
Fear & .073 & .840 \\
Happiness & .960 & .732 \\
Negative Surprise & .097 & .875 \\
Neutral & .469 & .184 \\
Positive Surprise & .875 & .875 \\
Sadness & .052 & .288 \\
\bottomrule
\end{tabular}}
\caption{Mapping from discrete labels to continuous valence and arousal values.}\label{tab:mapping}
\end{table}

\revision{
There are a few similar attempts to mapping discrete to continuous emotion models, but no
agreed-upon gold standard method to do so.
Our decision for this particular method is motivated by three criteria: 1) the method should yield a numeric value (in contrast to approaches like~\cite{Amiriparian24-EEH, gerczuk2021emonet} that utilize categories such as “low valence”  etc.) 2) the values should, of course, match our expectations based on Russel’s circumplex model~\cite{russell1980circumplex}
regarding the position of the discrete emotions in the V/A space, and 3) the method must be able to account for all labels in the dataset. We could, e.g., not utilize the V/A mappings for discrete emotions collected in~\cite{hoffmann2012mapping}, as they do not obtain values for surprise, disgust, and neutral. Admittedly, the method we selected has this problem for \textit{negative surprise} as well, but we found a relatively straightforward way to make up for this shortcoming.

We validate the mapping approach by obtaining additional valence/arousal labels from the same annotator for $3$ randomly selected stories. \Cref{tab:validation} reports the Pearson correlation between these direct valence/arousal annotations and those obtained by the proposed mapping. }

\begin{table}[h!]
    \centering\resizebox{.9\columnwidth}{!}{
    \begin{tabular}{llrr}
    \toprule
    Story (author) & $\rho$ Valence & $\rho$ Arousal \\ \midrule 
    \textit{emperor (Andersen)} & .7943 & .6151 \\
    \textit{hansel\_and\_gretel (Grimms)} & .8498 & .5765 \\
    \textit{the\_tale\_of\_jemima... (Potter)} & .7504 & .6088 \\
    \bottomrule
    \end{tabular}}\caption{Mapping approach validation on three stories. Reported are Pearson's correlations between direct V/A annotations and pseudo-V/A annotations as computed by the mapping from discrete labels.}\label{tab:validation}
\end{table}
\revision{
The correlations illustrate again that the difficulty of the problem varies for different stories. Moreover, the correlations for valence are higher than those for arousal, indicating that the method may capture valence better than arousal. This observation may also contribute to explaining why the automatic prediction of arousal proves to be more difficult than the prediction of valence.}

\section{Split Statistics}\label{appdx:split}

\Cref{tab:partition} displays detailed statistics for the split into training, development, and test sets.

\begin{figure}[h!]
    \centering
    \subfloat[Valence Values]{
    \includegraphics[width=.9\columnwidth]{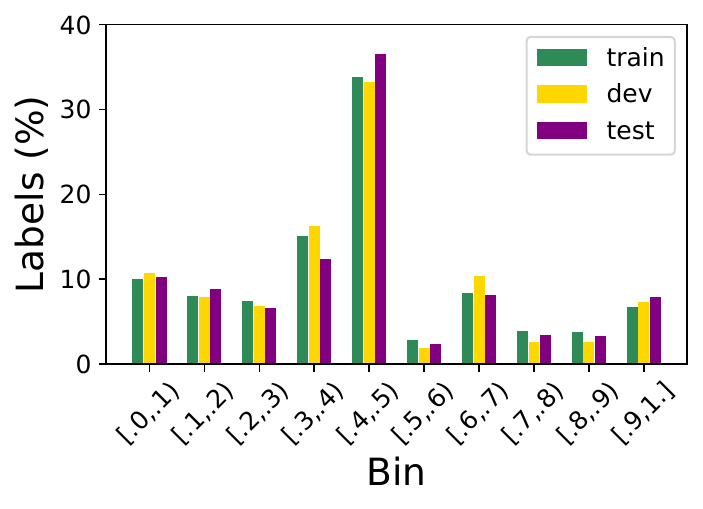}} \\
    \subfloat[Arousal Values]{
    \includegraphics[width=.9\columnwidth]{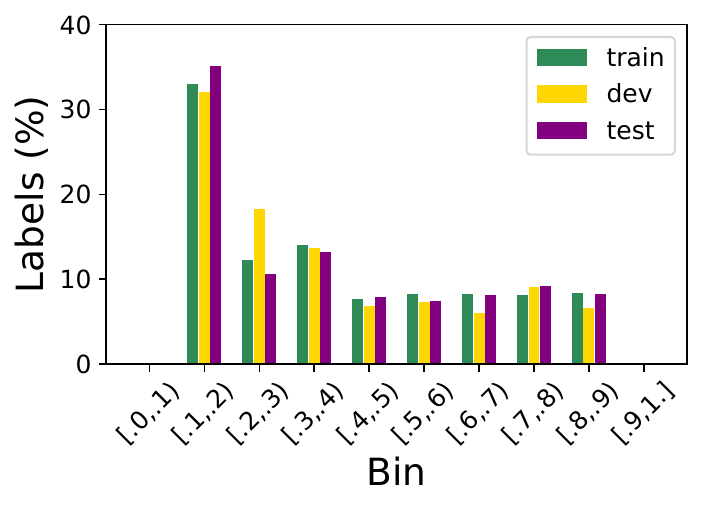}}
    \caption{Distributions of binned valence and arousal values in the created training, development (\textit{dev}), and test partitions.}
    \label{fig:split_bins}  
\end{figure}

\begin{table}[h!]
    \centering\resizebox{.8\columnwidth}{!}{
    \begin{tabular}{lrrrr}
    \toprule
         & \multicolumn{1}{c}{Overall} & \multicolumn{1}{c}{Grimm} & \multicolumn{1}{c}{HCA} & \multicolumn{1}{c}{Potter} \\ \midrule

         \multicolumn{5}{l}{\textbf{train}} \\ 
         
         stories & 118 & 54 (45.76\,\%) & 51 (43.22\,\%) & 13 (11.02\,\%) \\ 
         sentences & 10,121 & 3,621 (35.78\,\%) & 5,246 (51.38\,\%) & 1,254 (12.39\,\%) \\
         \midrule

\multicolumn{5}{l}{\textbf{development}} \\
         stories & 25 & 9 (36.00\,\%) & 13 (52.00\,\%) & 3 (12.00\,\%) \\
         sentences & 2,384 & 604 (25.34\,\%) & 1,494 (58.47\,\%) & 386 (16.19\,\%) \\
         \midrule
         
\multicolumn{5}{l}{\textbf{test}} \\
         stories & 26 & 14 (53.85\,\%) & 9 (34.62\,\%) & 3 (11.54\,\%) \\
         sentences & 2,379 & 1\,011 (42.50\,\%) & 1\,072 (45.06\,\%) & 296  (12.44\,\%) \\
         \bottomrule
    \end{tabular}}
    \caption{Dataset split statistics for every partition and author. For each author, the absolute number of stories as well as sentences in each partition is given. The percentage values denote the share of the author's stories/sentences in the stories/sentences of the respective partition.}
    \label{tab:partition}
\end{table}
\Cref{fig:split_bins} shows that the continuous label distributions are fairly similar in the different partitions.

\section{Further Experiment Details}\label{appdx:computation}
\revision{All experiments were carried out on an NVIDIA RTX3090 GPU and took about $200$ GPU hours in total. For illustration, we calculate the rough number of training/prediction steps for one experiment, \ie one configuration (\eg $C=4$) and one seed. The Alm dataset comprises about $15$k data points (about $10$k of which are used for training), the Gutenberg dataset contains about $100$k sentences. Assuming that steps 1) and 4) in Figure 4 run for 5 epochs each and step 3) takes one epoch, all of them using a batch size of 4, we end up with $(5*(100$k$ + 10$k$) + 110$k$) / 4 = 165$k training steps. Multiplying this with $5$ seeds and $5$ $C$-configurations, results in about $4$M training steps overall, notwithstanding some preliminary hyperparameter optimization and the additional \textit{author-independent} experiments. These rather large resource requirements also motivate our choice for the relatively small $304$M parameter DeBERTa model.}

\section{Gutenberg Corpus}\label{appdx:gb}

In \Cref{tab:gutenberg}, all books used for creating the \textsc{Gb} are listed.
We make sure not to include tales written by the three authors in the labeled dataset. We do not carry out any further filtering or manual screening steps. Basic preprocessing steps such as the removal of footnotes and images are conducted before we split the stories into sentences utilizing the PySBD~\cite{sadvilkar-neumann-2020-pysbd} library.

\section{Further Qualitative Analysis}\label{appdx:analysis}
In this section, we provide further examples of passages for which the model's predictions result in large errors, thus extending~\Cref{ssec:analysis_qualitative}. 

\Cref{tab:example_marriage} displays a text passage revolving around the theme of marriage. The model predicts high valence values, arguably due to this oftentimes positive topic. However, in this particular context, the planned marriage is viewed as negative by the protagonist.
\begin{table}[ht!]\centering
 \resizebox{.95\linewidth}{!}{

\begin{tabular}{lrr}
\toprule
\midrule 

\multicolumn{1}{l}{\textbf{Story}: \textit{li\_tiny} (Andersen)} \\

\multicolumn{3}{l}{\textbf{Context}: Protagonist (\textit{Tiny}) is about to get married against her will.
}\\

\midrule
\midrule

\multicolumn{1}{l}{\textbf{Sentence}} &  \multicolumn{1}{r}{\textbf{V pred}} & \multicolumn{1}{r}{\textbf{V GS}} \\ 
\midrule

\multirow{1}{250pt}{``You are going to be married, Tiny,'' said the field-mouse.}  & .8392 & .052 \\ 
\midrule

\multirow{1}{250pt}{"My neighbor has asked for you.} 
& .4902 & .469 \\
\midrule

\multirow{1}{250pt}{What good fortune for a poor child like you.} 
& .7692 & .469 \\

\midrule
\bottomrule
\end{tabular}}
\caption{Passage from \textit{li\_tiny} (Andersen) with valence predictions (\textit{pred}) and gold standard (\textit{GS}).}\label{tab:example_marriage}
\end{table}

In~\Cref{tab:example_fisherman}, an example is provided in which the model assesses a situation as positive, in which a person has gained a great amount of power. The gold standard, in contrast, assigns low valence values to this passage, as the protagonist exhibits greed and megalomania, aspects seemingly ignored by the model. 
\begin{table}[ht!]\centering
 \resizebox{.95\linewidth}{!}{

\begin{tabular}{lrr}
\toprule
\midrule 

\multicolumn{1}{l}{\textbf{Story}: \textit{the\_fisherman\_and\_his\_wife} (Grimms)} \\

\multicolumn{3}{l}{\textbf{Context}: The protagonist's wife (\textit{Ilsabill)} has acquired an absurd amount of power.
}\\

\midrule
\midrule

\multicolumn{1}{l}{\textbf{Sentence}} &  \multicolumn{1}{r}{\textbf{V pred}} & \multicolumn{1}{r}{\textbf{V GS}} \\ 
\midrule

\multirow{2}{250pt}{Then the fisherman went home, and found Ilsabill sitting on a throne that was two miles high.}  & .5495 & .200 \\ 
& & \\
\midrule

\multirow{2}{250pt}{And she had three great crowns on her head, and around her stood all the pomp and power of the Church.} 
& .6597 & .357 \\
& & \\
\midrule

\multirow{4}{250pt}{And on each side of her were two rows of burning lights, of all sizes, the greatest as large as the highest and biggest tower in the world, and the least no larger than a small rushlight.} 
& .6668 & .357 \\
& & \\
& & \\
& & \\
\midrule

\midrule
\bottomrule
\end{tabular}}
\caption{Passage from \textit{the\_fisherman\_and\_his\_wife} (Grimms) with valence predictions (\textit{pred}) and gold standard (\textit{GS}).}\label{tab:example_fisherman}
\end{table}

The phenomenon of our model missing out on the overall tone of stories is further exemplified by the text in~\Cref{tab:example_pigs}. Here, the protagonists are pigs behaving like and interacting with humans, which gives the entire story a funny mood. In this context, a typically rather exciting situation (being interrogated by the police) is not assigned a high arousal value by the gold standard -- different from the model.
\begin{table}[ht!]\centering
 \resizebox{.95\linewidth}{!}{

\begin{tabular}{lrr}
\toprule
\midrule 

\multicolumn{1}{l}{\textbf{Story}: \textit{the\_tale\_of\_pigling\_bland} (Potter)} \\

\multicolumn{3}{l}{\textbf{Context}: Pigs are interrogated by a policeman.
}\\

\midrule
\midrule

\multicolumn{1}{l}{\textbf{Sentence}} &  \multicolumn{1}{r}{\textbf{A pred}} & \multicolumn{1}{r}{\textbf{A GS}} \\ 
\midrule

\multirow{1}{250pt}{What's that, young Sirs?}  & .5057 & .184 \\ 
\midrule

\multirow{1}{250pt}{Stole a pig?} 
& .5014 & .184 \\
\midrule

\multirow{1}{250pt}{Where are your licenses? said the policeman.} 
& .6877 & .184 \\

\midrule
\bottomrule
\end{tabular}}
\caption{Passage from \textit{the\_tale\_of...} (Potter) with arousal predictions (\textit{pred}) and gold standard (\textit{GS}).}\label{tab:example_pigs}
\end{table}

Another example from an overall funny story is given in~\Cref{tab:example_fred}. The protagonists encounter several robbers, but the situation is labeled with a \textit{neutral} valence value in the gold standard. The model assigns low valence values, missing out on the funny tone of the entire story.
\begin{table}[ht!]\centering
 \resizebox{.95\linewidth}{!}{

\begin{tabular}{lrr}
\toprule
\midrule 

\multicolumn{1}{l}{\textbf{Story}: \textit{frederick\_and\_catherine} (Grimms)} \\

\multicolumn{3}{l}{\textbf{Context}: Protagonists (\textit{Frederick and Catherine}) encounter robbers.
}\\

\midrule
\midrule

\multicolumn{1}{l}{\textbf{Sentence}} &  \multicolumn{1}{r}{\textbf{V pred}} & \multicolumn{1}{r}{\textbf{V GS}} \\ 
\midrule

\multirow{2}{250pt}{Scarcely were they up, than who should come by but the very rogues they were looking for.}  & .2310 & .469 \\ 
& & \\
\midrule

\multirow{4}{250pt}{They were in truth great rascals, and belonged to that class of people who find things before they are lost; they were tired; so they sat down and made a fire under the very tree where Frederick and Catherine were.} 
& .2891 & .469 \\
& & \\
& & \\
& & \\
\midrule

\midrule
\bottomrule
\end{tabular}}
\caption{Passage from \textit{frederick\_and\_catherine} (Grimms) with valence predictions (\textit{pred}) and gold standard (\textit{GS}).}\label{tab:example_fred}
\end{table}

Moreover, we provide further examples of the model struggling with stories within stories. In the passage given in~\Cref{tab:example_fire}, the model overestimates the arousal value, as the story told about a great fire is arguably very exciting.
\begin{table}[ht!]\centering
 \resizebox{.95\linewidth}{!}{

\begin{tabular}{lrr}
\toprule
\midrule 

\multicolumn{1}{l}{\textbf{Story}: \textit{a\_story} (Andersen)} \\

\multicolumn{3}{l}{\textbf{Context}: A man recounts a memory where he witnessed a fire.
}\\

\midrule
\midrule

\multicolumn{1}{l}{\textbf{Sentence}} &  \multicolumn{1}{r}{\textbf{A pred}} & \multicolumn{1}{r}{\textbf{A GS}} \\ 
\midrule

\multirow{2}{250pt}{All burnt down- a great heat rose, such as sometimes overcomes me.}  & .6779 & .460 \\ 
& & \\
\midrule

\multirow{7}{250pt}{I myself helped to rescue cattle and things, nothing alive burnt, except a flight of pigeons, which flew into the fire, and the yard dog, of which I had not thought; one could hear him howl out of the fire, and this howling I still hear when I wish to sleep; and when I have fallen asleep, the great rough dog comes and places himself upon me, and howls, presses, and tortures me.} 
& .7914 & .288 \\
& & \\
& & \\
& & \\
& & \\
& & \\
& & \\
\midrule

\midrule
\bottomrule
\end{tabular}}
\caption{Passage from \textit{a\_story} (Andersen) with arousal predictions (\textit{pred}) and gold standard (\textit{GS}).}\label{tab:example_fire}
\end{table}

The example presented in~\Cref{tab:example_oldbach} demonstrates a passage where valence is overestimated by the model. Here, happy memories are recalled in a sad context, giving the text a sad mood that is not properly assessed by the model.
\begin{table}[ht!]\centering
 \resizebox{.95\linewidth}{!}{

\begin{tabular}{lrr}
\toprule
\midrule 

\multicolumn{1}{l}{\textbf{Story}: \textit{old\_bach} (Andersen)} \\

\multicolumn{3}{l}{\textbf{Context}: a sad old bachelor recalls his youth.
}\\

\midrule
\midrule

\multicolumn{1}{l}{\textbf{Sentence}} &  \multicolumn{1}{r}{\textbf{V pred}} & \multicolumn{1}{r}{\textbf{V GS}} \\ 
\midrule

\multirow{2}{250pt}{How much came back to his remembrance as he looked through the tears once more on his native town!}  & .8102 & .469 \\ 
& & \\
\midrule

\multirow{6}{250pt}{The old house was still standing as in olden times, but the garden had been greatly altered; a pathway led through a portion of the ground, and outside the garden, and beyond the path, stood the old apple-tree, which he had not broken down, although he talked of doing so in his trouble.} 
& .7996 & .469 \\
& & \\
& & \\
& & \\
& & \\
& & \\
\midrule

\multirow{4}{250pt}{The sun still threw its rays upon the tree, and the refreshing dew fell upon it as of old; and it was so overloaded with fruit that the branches bent towards the earth with the weight.} 
& .9574 & .634 \\
& & \\
& & \\
& & \\

\midrule
\bottomrule
\end{tabular}}
\caption{Passage from \textit{old\_bach} (Andersen) with valence predictions (\textit{pred}) and gold standard (\textit{GS}).}\label{tab:example_oldbach}
\end{table}

\begin{table*}[h!]\centering
 \resizebox{.95\linewidth}{!}{
\begin{tabular}{rl}
\toprule 
\multicolumn{1}{l}{Gutenberg ID} & Book Title \\
\midrule
22656 & Red Cap Tales, Stolen from the Treasure Chest of the Wizard of the North \\ 
19713 & The Laughing Prince: Jugoslav Folk and Fairy Tales\\ 
4357 & American Fairy Tales \\ 
19207 & The Firelight Fairy Book \\ 
17034 & English Fairy Tales\\ 
24714 & Fairy Tales from Brazil: How and Why Tales from Brazilian Folk-Lore \\ 
20748 & Favorite Fairy Tales \\ 
20366 & Wonderwings and other Fairy Stories \\ 
7439 & English Fairy Tales \\ 
24593 & The Oriental Story Book: A Collection of Tales \\ 
22420 & The Book of Nature Myths  \\ 
19734 & The Fairy Book \\ 
8599 & Fairy Tales from the Arabian Nights  \\ 
9368 & Welsh Fairy Tales \\ 
16537 & Myths That Every Child Should Know \\ 
24473 & The Cat and the Mouse: A Book of Persian Fairy Tales \\ 
22168 & The golden spears, and other fairy tales \\ 
25502 & Hero-Myths \& Legends of the British Race\\ 
22175 & Stories from the Ballads, Told to the Children \\ 
24737 & The Children of Odin: The Book of Northern Myths\\ 
22693 & A Book of Myths \\ 
677 & The Heroes; Or, Greek Fairy Tales for My Children\\ 
23462 & More Russian Picture Tales\\ 
15145 & My Book of Favourite Fairy Tales \\ 
4018 & Japanese Fairy Tales \\ 
11319 & The Fairy Godmothers and Other Tales \\ 
15164 & Folk Tales Every Child Should Know \\ 
7871 & Dutch Fairy Tales for Young Folks\\ 
7488 & Celtic Tales, Told to the Children \\ 
14916 & Fairy Tales Every Child Should Know \\ 
20552 & Roumanian Fairy Tales \\ 
2892 & Irish Fairy Tales \\ 
7885 & Celtic Fairy Tales \\ 
22096 & Stories the Iroquois Tell Their Children \\ 
25555 & Fairy Tales of the Slav Peasants and Herdsmen \\ 
14421 & Wilson's Tales of the Borders and of Scotland, Volume 24 \\ 
7128 & Indian Fairy Tales\\ 
6622 & Legends That Every Child Should Know; a Selection of the Great Legends of All...\\ 
8675 & Welsh Fairy-Tales and Other Stories\\ 
22373 & Russian Fairy Tales: A Choice Collection of Muscovite Folk-lore \\ 
22886 & Cinderella in the South: Twenty-Five South African Tales\\ 
22248 & The Indian Fairy Book: From the Original Legends\\ 
24811 & Viking Tales\\ 
18674 & A Chinese Wonder Book \\ 
22396 & King Arthur's Knights \\

\bottomrule
\end{tabular}}
\caption{List of books in the Gutenberg corpus.}\label{tab:gutenberg}
\end{table*}

\begin{acronym}
\acro{A}[A]{Arousal}
\acro{ABC}[ABC]{Airplane Behaviour Corpus}
\acro{AD}[AD]{Anger Detection}
\acro{AFEW}[AFEW]{Acted Facial Expression in the Wild)}
\acro{AI}[AI]{Artificial Intelligence}
\acro{ANN}[ANN]{Artificial Neural Network}
\acro{ASO}[ASO]{Almost Stochastic Order}
\acro{ASR}[ASR]{Automatic Speech Recognition}

\acro{BN}[BN]{batch normalisation}
\acro{BiLSTM}[BiLSTM]{Bidirectional Long Short-Term Memory}
\acro{BES}[BES]{Burmese Emotional Speech}
\acro{BoAW}[BoAW]{Bag-of-Audio-Words}
\acro{BoDF}[BoDF]{Bag-of-Deep-Feature}
\acro{BoW}[BoW]{Bag-of-Words}

\acro{CASIA}[CASIA]{Speech Emotion Database of the Institute of Automation of the Chinese Academy of Sciences}
\acro{CCC}[CCC]{Concordance Correlation Coefficient}
\acro{CVE}[CVE]{Chinese Vocal Emotions}
\acro{CNN}[CNN]{Con\-vo\-lu\-tion\-al Neural Network}
\acro{CRF}[CRF]{Conditional Random Field}
\acro{CRNN}[CRNN]{Con\-vo\-lu\-tion\-al Recurrent Neural Network}

\acro{DEMoS}[DEMoS]{Database of Elicited Mood in Speech}
\acro{DES}[DES]{Danish Emotional Speech}
\acro{DENS}[DENS]{Dataset for Emotions of Narrative Sequences}
\acro{DNN}[DNN]{Deep Neural Network}
\acro{DS}[DS]{\ds}

\acro{eGeMAPS}[eGeMAPS]{extended version of the Geneva Minimalistic Acoustic Parameter Set}
\acro{EMO-DB}[EMO-DB]{Berlin Database of Emotional Speech}
\acro{EmotiW}[EmotiW 2014]{Emotion in the Wild 2014}
\acro{eNTERFACE}[eNTERFACE]{eNTERFACE'05 Audio-Visual Emotion Database}
\acro{EU-EmoSS}[EU-EmoSS]{EU Emotion Stimulus Set}
\acro{EU-EV}[EU-EV]{EU-Emotion Voice Database}
\acro{EWE}[EWE]{Evaluator-Weighted Estimator}

\acro{AIBO}[FAU Aibo]{FAU Aibo Emotion Corpus}
\acro{FCN}[FCN]{Fully Convolutional Network}
\acro{FFT}[FFT]{fast Fourier transform}

\acro{GAN}[GAN]{Generative Adversarial Network}
\acro{GEMEP}[GEMEP]{Geneva Multimodal Emotion Portrayal}
\acro{GRU}[GRU]{Gated Recurrent Unit}
\acro{GVEESS}[GVEESS]{Geneva Vocal Emotion Expression Stimulus Set}

\acro{IEMOCAP}[IEMOCAP]{Interactive Emotional Dyadic Motion Capture}

\acro{LDA}[LDA]{Latent Dirichlet Allocation}
\acro{LSTM}[LSTM]{Long Short-Term Memory}
\acro{LLD}[LLD]{low-level descriptor}
\acro{LLM}[LLM]{Large Language Model}

\acro{MELD}[MELD]{Multimodal EmotionLines Dataset}
\acro{MES}[MES]{Mandarin Emotional Speech}
\acro{MFCC}[MFCC]{Mel-Frequency Cepstral Coefficient}
\acro{MSE}[MSE]{Mean Squared Error}
\acro{MIP}[MIP]{Mood Induction Procedure}
\acro{MLP}[MLP]{Multilayer Perceptron}
\acro{NLP}[NLP]{Natural Language Processing}
\acro{NLU}[NLU]{Natural Language Understanding}
\acro{NMF}[NMF]{Non-negative Matrix Factorization}

\acro{ReLU}[ReLU]{Rectified Linear Unit}
\acro{REMAN}[REMAN]{Relational EMotion ANnotation}
\acro{RMSE}[RMSE]{root mean square error}
\acro{RNN}[RNN]{Recurrent Neural Network}
\acrodefplural{RNN}[RNNs]{Recurrent Neural Networks}

\acro{SER}[SER]{Speech Emotion Recognition}
\acro{SGD}[SGD]{Stochastic Gradient Descent}
\acro{SVM}[SVM]{Support Vector Machine}
\acro{SIMIS}[SIMIS]{Speech in Minimal Invasive Surgery}
\acro{SmartKom}[SmartKom]{SmartKom Multimodal Corpus}
\acro{SEND}[SEND]{Stanford Emotional Narratives Dataset}
\acro{SUSAS}[SUSAS]{Speech Under Simulated and Actual Stress}

\acro{TER}[TER]{Textual Emotion Recognition}
\acro{TTS}[TTS]{Text-to-Speech}

\acro{UAR}[UAR]{Unweighted Average Recall}
\acro{V}[V]{Valence}
\acro{VRNN}[VRNN]{Variational Recurrent Neural Networks}
\acro{WSJ}[WSJ]{Wall Street Journal}
\end{acronym}
\end{document}